\UseRawInputEncoding
\documentclass[runningheads]{llncs}
\usepackage{graphicx}
\usepackage{makecell}
\usepackage{cite}
\usepackage{amsmath,amssymb,amsfonts}
\usepackage{algorithmicx,algorithm}
\usepackage{graphicx}
\usepackage{textcomp}
\usepackage{xcolor}
\usepackage{booktabs}
\usepackage{array}
\usepackage[noend]{algpseudocode}
\usepackage{subcaption}
\usepackage{multirow}
\usepackage{breakurl}

\begin{document}
\title{Learning to Remove: Towards Isotropic \\
	Pre-trained BERT Embedding}
%
%
\author{Yuxin Liang\inst{1} \and Rui Cao\inst{1} \and
Jie Zheng\thanks{Corresponding author.}\inst{1} \and Jie Ren\inst{2} \and Ling Gao\inst{1}}
\authorrunning{Y. Liang et al.}
%
\institute{Northwest University, Xi'an, China \\
\email{\{liangyuxin,caorui\}@stumail.nwu.edu.cn, \{jzheng,gl\}@nwu.edu.cn} \and
Shannxi Normal University, Xi'an, China\\
\email{renjie@snnu.edu.cn}}
\maketitle              
\begin{abstract}
Research in word representation shows that isotropic embeddings can significantly improve performance on downstream tasks. However, we measure and analyze the geometry of pre-trained BERT embedding and find that it is far from isotropic.  We find that the word vectors are not centered around the origin, and the average cosine similarity between two random words is much higher than zero, which indicates that the word vectors are distributed in a narrow cone and deteriorate the representation capacity of word embedding. 
We propose a simple, and yet effective method to fix this problem: remove several dominant directions of BERT embedding with a set of learnable weights. We train the weights on word similarity tasks and show that processed embedding is more isotropic. Our method is evaluated on three standardized tasks: word similarity, word analogy, and semantic textual similarity. In all tasks, the word embedding processed by our method consistently outperforms the original embedding (with average improvement of 13\% on word analogy and 16\% on semantic textual similarity) and two baseline methods. Our method is also proven to be more robust to changes of hyperparameter.

\end{abstract}
\section{Introduction}
With the rise of Transformers\cite{vaswani2017attention}, its derivative model BERT\cite{devlin2018bert} stormed the NLP field due to its excellent performance in various tasks. A new word BERTology has been defined to describe the related research work carried out around BERT\cite{rogers2020primer}. Our work is focused on BERT input embedding that is both critical to the BERT model and easy to overlook.

From static word embeddings to contextual embeddings, the word embedding technology is constantly evolving, but it is still not perfect so far. The analysis from \cite{mu2018all} implies that in static embeddings, all word embeddings share a common vector and have several dominant directions. These anisotropic geometric properties strongly affect the representation of words. Furthermore \cite{hasan2017word,gong2018frage,ethayarajh2019contextual,zhou2019getting}, it was found that the word embeddings of contextual language models also have those anisotropic geometric properties, and the main reason for this phenomenon is the large number of low-frequency words in the corpus. The embedding of most words in the vocabulary is pushed into a similar direction that is negatively correlated with most hidden states, thus clusters in a localized region of the embedding space. This is known as the representation degeneration problem \cite{gao2018representation}. 

Does BERT embedding also share this problem? The answer is Yes: pre-trained BERT embedding also suffers from strong anisotropy, meaning the average cosine similarity value is significantly higher than zero, and word vectors clustering in narrow cones in the vector space. This phenomenon can result in a word representation have a high similarity to an unrelated word, affecting the expressive power. In addition, the anisotropic property can also affect the accuracy of downstream tasks \cite{gong2018frage,liu2019unsupervised}. Therefore, various attempts have been made to eliminate such a property so that the learned word vectors are more divergent and distinguishable in the Euclidean space.

Referring to the all-but-the-top method\cite{mu2018all} (denoted as ABTT for short), they remove the common vector along with several dominant directions computed by PCA (principal component analysis) in word vectors. After applying this method to BERT embedding, we find that the ABTT method could improve the geometric properties and task performance of BERT embedding. However, as the number of selected dominant directions increases, the effectiveness of the ABTT method decreases gradually, as shown in section 5. This is intuitive since some useful linguistic information contained within these directions is inevitably lost as the number of removed directions increases.

We propose a weighted removal method that sets a learnable weight for each dominant direction, dynamically adjusting the ratio of removal. We train on the word similarity task to obtain the weight for each dominant direction. Our approach is more flexible and performs better on the evaluation task compared to removing the dominant directions directly and completely. Our method alleviates representation degradation and performs more stable in comparative experiments on the word similarity, word analogy, and text similarity tasks. The code is available here\footnote[1]{https://github.com/liangyuxin42/weighted-removal}.

Our contributions are as follows:

1. We measure, visualize and analyze the geometry of pre-trained BERT embedding. We find that the pre-trained BERT embedding is anisotropic and the norm/average cosine similarity of word vectors has a strong correlation with word frequency. Further, we provide some intuitive explanation and theoretical analysis for the above phenomena.

2. We propose a weighted removal method that learns to remove the dominant directions. The key point of our approach is using a set of learnable weights trained on word similarity tasks to decide the proportions of the removal directions.

3. We evaluate our method on three tasks: word similarity, word analogy, and semantic textual similarity. We compare the performance of our method with three baselines: original pre-trained BERT embedding, ABTT method, and conceptor negation method (denoted as CN for short). Our method outperforms three baselines in most cases and maintains relatively stable performance as hyperparameter changes. We also analyze the impact of our method on the geometry of word embedding and conclude that our method makes word embedding more isotropic and expressive.

\section{Related Work} 
The BERT model is introduced in \cite{devlin2018bert}, a language model based on Transformer (containing 12-layer to 24-layer Transformer encoders), pre-trained on a hybrid corpus. BERT's internal operation includes first embed the tokens by a pre-trained embedding layer and combine with position and segment information. These initial embeddings run through several transformer encoder layers to produce the contextual embedding for the current task. Our study is based on the pre-trained BERT embedding and focuses on its geometric properties.

Mu et al.\cite{mu2018all} explore the anisotropic geometry of static word embeddings, i.e., the word vectors have a non-zero mean and most word vectors have several dominant directions. They propose to remove the common vector and dominant directions from the word embedding to capture stronger linguistic regularities. There are other attempts to fix the anisotropic geometry of static word embeddings: Liu et al.\cite{liu2019unsupervised} use conceptors to suppress those latent features of word vectors with high variances. Hasan et al.\cite{hasan2017word} re-embed pre-trained word embeddings with a stage of manifold learning. Zhou et al.\cite{zhou2019getting} focuses on linear alignment of word embeddings and find that aligning with an isotropic noise can deliver better results. 

Ethayarajh et al.\cite{ethayarajh2019contextual} find that the contextualized representations are not isotropic in layers of contextualizing models such as BERT and GPT-2. Gong et al.\cite{gong2018frage} find that contextual word embeddings are biased towards word frequency and use adversarial training to learn a frequency-agnostic word representation. 

Gao et al.\cite{gao2018representation} focus on the learned word embeddings of natural language generation model training through likelihood maximization and find that the word embedding tends to degenerate and is distributed into a narrow cone in vector space which limits its representation power. They propose a regularization method that minimizes the cosine similarities between any word vector pair to alleviate this problem. Wang et al.\cite{wang2020improving} propose a spectrum control method that guides the spectra training of the output embedding with a slow-decaying singular value prior distribution to tackle the representation degeneration problem. Other researches also touch on the geometry of contextual word embedding, such as Reif et al.\cite{reif2019visualizing}, Karve et al.\cite{karve2019conceptor}, Zhou et al.\cite{zhou2020isobn}.

The above word embedding post-processing methods are based only on the geometric features, while we make the correction semantically more reasonable by introducing information of word similarity. The methods of regularizing word similarity during training require retraining the model. In contrast, our post-processing method is more computationally efficient.

The main difference between our method and ABTT is that we add learnable weights to each dominant direction and train the weights on the word similarity task. Furthermore, We apply our method on pre-trained BERT embeddings to fix the representation degeneration problem. We compare the performance of our method with ABTT and CN\cite{liu2019unsupervised} which is a successor of the ABTT method.

\section{Observation} 
In this section, we illustrate the anisotropic geometry of pre-trained BERT embedding and provide some explanation for those phenomena. Our study is based on pre-trained bert-base-uncased and bert-large-uncased models provided by huggingface\cite{wolf2019huggingface}.
\subsection{Anisotropic geometry of pre-trained BERT embedding}

\begin{table}[t]
	\caption{Basic geometric information of pre-trained BERT embedding (BERT-base-uncased and BERT-large-uncased).}
	\label{tab1}
	\centering
	\begin{tabular}{|l|l|l|l|}
		\hline
		Embedding  & Average Vector Length & Vector Average Length & Average Cosine Similarity \\
		\hline
		BERT-base  & \makecell[c]{0.939}  & \makecell[c]{1.401}   & \makecell[c]{0.444} \\
		BERT-large & \makecell[c]{0.800}  & \makecell[c]{1.453}   & \makecell[c]{0.299} \\
		\hline
	\end{tabular}
\end{table}

Let $v(w)\in\mathbb{R}^e$ be the embedding of a token $w$ in a vocabulary $V$, and $E$ of shape $(|V|, e)$ be the embedding matrix. We observe the following phenomena in the pre-trained BERT embedding.

\textbf{Non-zero mean}: The average vector $\mu = \frac{1}{|V|} \Sigma_{w \in V} v(w)$ is not centered around the origin. In fact, this common vector $\mu$ occupies a large proportion of all $v(w)$. The norm of $\mu$ is more than 1/2 of the average norm of all $v(w)$ (Table \ref{tab1}).

\textbf{Non-zero average cosine similarity}: The average cosine similarity is calculated by $$cos_{avg} = \frac{\Sigma_{w_i \in V} \Sigma_{w_j \in V} cosine(v(w_i),v(w_j))}{|V|^2}.$$ As shown in Table \ref{tab1}, the average cosine similarity is much higher than zero, which means instead of uniformly distributed in vector space as one expects from an expressive word embedding, the word vectors are distributed into a narrow cone.

\begin{figure}[t]
	\centering
	\label{fig1}
	\includegraphics[width=0.5\linewidth]{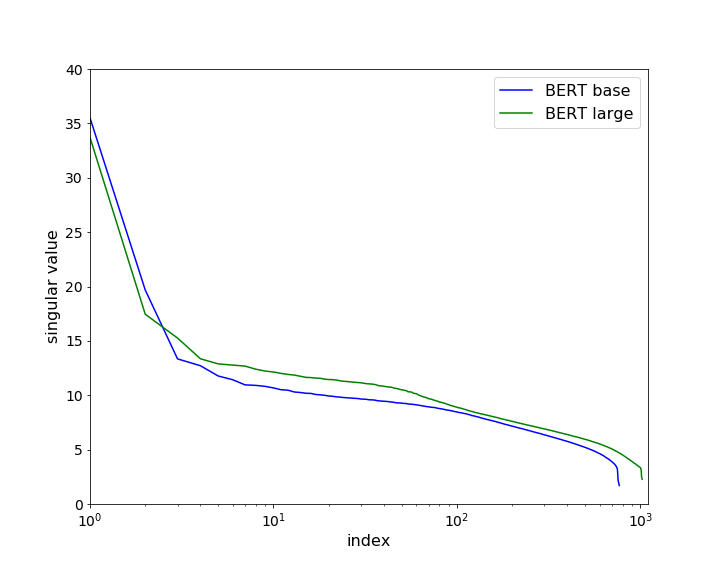}
	\caption{Singular values of embedding matrix. The singular values of BERT embedding matrix decay exponentially. Note that the coordinates are semi-logarithmic.}
\end{figure}

\textbf{Fast singular value decay}: We reparameterize the embedding matrix $E$ by singular value decomposition (SVD) and obtain the singular values of matrix $E$. As shown in Figure 1, the singular values of matrix $E$ decay exponentially.

\begin{figure}[t]
	\centering
	\setlength{\lineskip}{\medskipamount}
	\subcaptionbox{\label{pic2_a}}{\includegraphics[width=0.25\linewidth]{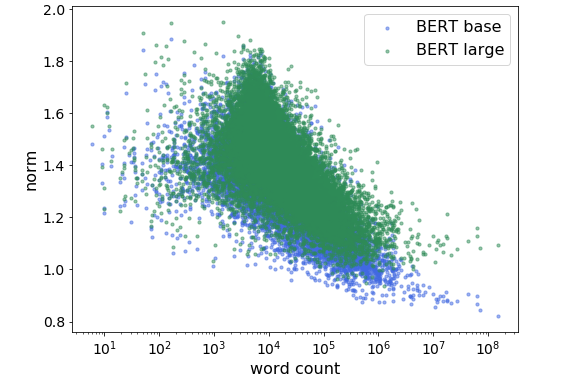}}\hfill
	\subcaptionbox{\label{pic2_b}}{\includegraphics[width=0.25\linewidth]{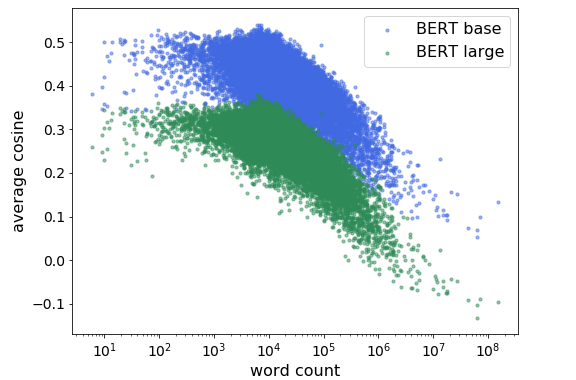}}\hfill
	\subcaptionbox{\label{pic2_c}}{\includegraphics[width=0.25\linewidth]{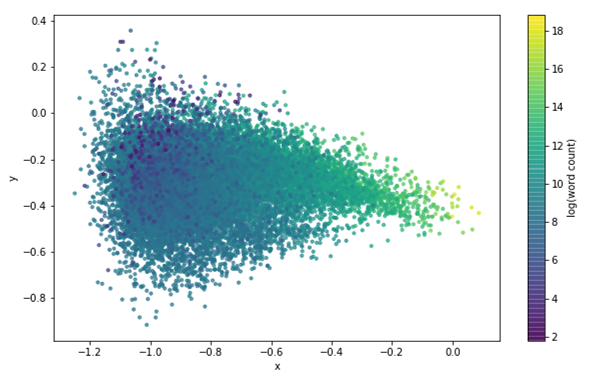}}\hfill
	\subcaptionbox{\label{pic2_d}}{\includegraphics[width=0.25\linewidth]{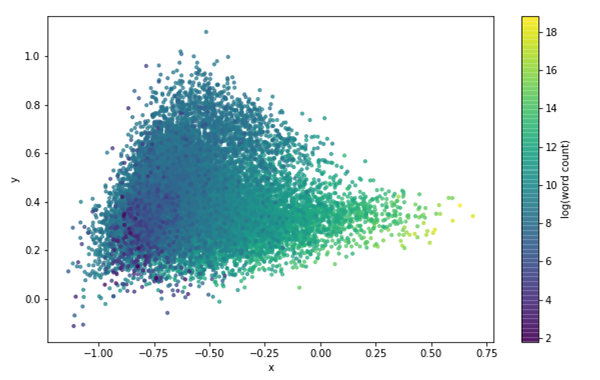}}\hfill
	\caption{Correlation between (a) vector norm and word frequency (b) average cosine similarity and word frequency in BERT base/large embedding. Projecting (c) BERT-base and (d) BERT-large embedding onto top two PCA directions.} \label{pic2}
\end{figure}

\textbf{Correlation with word frequency}: When we look into the relationship between the norm of $v(w)$, the average cosine similarity and the word frequency of word $w$, we notice an obvious correlation between them, as shown in Figure \ref{pic2_a}, \ref{pic2_b}. The Pearson correlation between norm/average cosine similarity and the logarithm of word count is about $-0.7$ which also shows a strong negative correlation.

We also visualize BERT embedding by projecting onto the top two PCA directions and use color to indicate word frequency, as shown in Figure \ref{pic2_c}, \ref{pic2_d}. We find that the first PCA coefficient encodes the word frequency to a significant degree with a high Pearson correlation (about $-0.7$).

The above phenomena reveal an image of the distribution of words with different frequencies in the embedding space. The words with high frequency are close to the origin (smaller norm) and distributed relatively uniform (lower average cosine similarity), and the words with low frequency are squeezed into a narrower cone and push away from the origin. This can lead to two words that are explicitly dissimilar to each other but whose corresponding word vectors may produce a high degree of similarity in the Euclidean space, thus affecting the performance of downstream tasks.

\subsection{Insights}

The above phenomena are very similar to the representation degeneration problem mentioned in \cite{gao2018representation}, inspired by their work, we provide some intuitive explanation and theoretical analysis for the above phenomena. 

In the pre-train process of BERT, the final hidden vectors corresponding to the mask tokens are fed into an output softmax over the vocabulary, as in a standard language model\cite{devlin2018bert}. Intuitively, during the training process, for any given hidden state, the embedding of the corresponding masked word will be pushed towards the direction of the hidden state to get a larger likelihood, while the embeddings of all other words will be pushed towards the negative direction of the hidden state to get a smaller likelihood. The words with lower frequency would be pushed more times towards the negative direction of all kinds of hidden states. As a result, low frequency words are squeezed into a narrower cone.

From the theoretical perspective, for any token $w_i$, its loss function can be divided into two pieces: piece $A_{w_i}$ for the part of training corpus that does not contain token $w_i$ in context, piece $B_{w_i}$ for the part containing token $w_i$. Let $P(context \in A_{w_i})$ and $P(context \in B_{w_i})$ denote the probability of a context belonging to piece A/B and $L_{A_{w_i}}(w_i)$ and $L_{B_{w_i}}(w_i)$ be the loss of piece A/B respectively. So the loss function of $w_i$ can be defined as:
	$$L_{w_i} = P(context \in A_{w_i})L_{A_{w_i}}(w_i) +P(context \in B_{w_i})L_{B_{w_i}}(w_i)$$

According to \cite{gao2018representation}, $L_{A_{w_i}}(w_i)$ is convex and would optimize embedding $w_i$ towards any uniformly negative direction of the hidden state in piece $A_{w_i}$ to infinity. $L_{B_{w_i}}(w_i)$ is complicated but when $P(context \in A_{w_i})$ is much larger than $ P(context \in B_{w_i})$, the optimal solution of $L_{w_i}$ is close to $L_{A_{w_i}}(w_i)$. A more detailed proof can be found in \cite{gao2018representation}.

With the interpretation above, we can explain the phenomena observed. For most words, $P(context \in A_{w_i})$ is much larger than $P(context \in B_{w_i})$, and piece A of different words has large overlaps. So most words will be optimized toward a similar direction in vector space, hence the average vector $\mu$ would also be in that direction and the average cosine similarity would be larger than zero.

For a low-frequency word, $P(context \in A_{w_i})$ is even larger. So a low-frequency word is affected more by the optimization of $L_{A_{w_i}}(w_i)$, therefore the word vectors of low-frequency words are likely to be close to each other and pushed farther away from the origin, which is consistent with our observation.

\section{Method}
In this section, we propose a simple but effective way to change the anisotropic geometry of pre-trained BERT embedding into a more isotropic and expressive one. According to the analysis in the previous section, the word embeddings have some dominant directions with a disproportional impact on word embedding. But removing these directions directly and completely may lose semantic and syntactic information within them. 

The core idea of our method is to remove several dominating directions causing the anisotropic problem with a learnable weight to each direction. We use PCA to determine the dominating directions. We denote our method as weighted removal (WR) algorithm for simplicity.

The weighted removal (WR) method includes three main steps: compute the dominant directions using PCA, learn a set of weights to remove those directions through a certain learning process, and remove directions with corresponding weights from word embedding. We formally achieve our method as Algorithm \ref{alg1}.

\begin{algorithm}[htbp]
	\caption{Weighted Removal (WR) Algorithm} 
	\label{alg1}
		\hspace*{0.02in} {\bf Input:} Word Embedding ${v(w)}\in\mathbb{R}^e$; $d$: number of directions being removed\\
		\hspace*{0.02in} {\bf Output:} 
		Processed word embedding $v^{'}(w)$
	\begin{algorithmic}[1]
		\State Compute the PCA components: 
		$u_1,..., u_{e} \leftarrow PCA({v(w), w \in V})$
		\State Acquire $d$ weights for $d$ directions through a learning process:
		$\alpha_1, ..., \alpha_d$
		\State Remove top $d$ dominating directions: 
		$v^{'}(w) \leftarrow v(w) - \sum_{i=1}^{d} \alpha_i (u^T_i v(w))u_i$
		\State \Return $v^{'}(w)$
	\end{algorithmic}
\end{algorithm}

\textbf{Learning process}: We choose word similarity as the training task. 
The unreasonable word similarity is the direct manifestation of the representation degeneration problem. Through adjusting word similarity, our method indirectly adjusts the geometric properties of the word vector distribution in vector space towards isotropic.

Giving two words $w_1$ and $w_2$ with their corresponding ground truth of similarity $S_{target}$, the prediction of word similarity is calculated as $S_{pred} = \hat{v}^{'}(w_1)^T \hat{v}^{'}(w_2) $. For simplicity, we denote the normalized direction of $v(w)$ as $\hat{v}(w)$, $\hat{v}(w) = \frac{v(w)}{\| v(w) \|}$.
The loss function is defined as: $L = MSELoss(S_{pred}, S_{target})$. After the training process, we get the $d$ adjusted weights for removing top $d$ dominating directions.

\textbf{Comparison Algorithm}: Our method is inspired by the ABTT method. We also apply the ABTT method and its direct successor method, the conceptor negation(CN) method proposed in \cite{liu2019unsupervised} on BERT embedding for comparison.

The ABTT algorithm directly removes the first $d$ dominant direction completely. The CN algorithm uses conceptors to suppress those latent features with high variances in the word embedding. We skip the step of removing the mean vector in the original ABTT algorithm because the word embedding processed by the weighted removal algorithm is naturally centered around the origin.

\section{Experiment}

Our experiment is based on the pre-trained BERT model provided by huggingface Transformers library. 
As for training datasets, we collect 8 commonly used word similarity datasets: RG65, WordSim-353 (WS), the rare-words (RW), the MEN, the MTurk-278\&MTurk-771, the SimLex-999, and the SimVerb-3500 dataset. We scale the human annotations into [-1, 1] to be consistent with cosine similarity calculation.

The number of dominating directions to be removed, $d$, is a hyper-parameter needs to be tuned. We choose a series of numbers from 1 to 200 as $d$ to perform the ABTT algorithm and our WR algorithm.
The performance of our algorithm is compared with 3 baseline: 

1. The original pre-trained BERT base/large uncased embedding (ORIG). 

2. Embeddings processed by ABTT algorithm. This algorithm also has the same hyper-parameter $d$. 

3. Embeddings processed by conceptor negation (CN) algorithm with the same setting as the original paper.

\subsection{Towards Expressive}
\textbf{Word Similarity}: For this experiment, we randomly separate the collection of word similarity datasets into 70\% for training and 30\% for testing. Given a pair of words, we calculate their similarity by cosine similarity. The word similarity task is evaluated in terms of Pearson's correlation with the human annotations.

\begin{figure}[t]
	\centering
	\setlength{\lineskip}{\medskipamount}
	\subcaptionbox{\label{fig3_a}}{\includegraphics[width=0.32\linewidth]{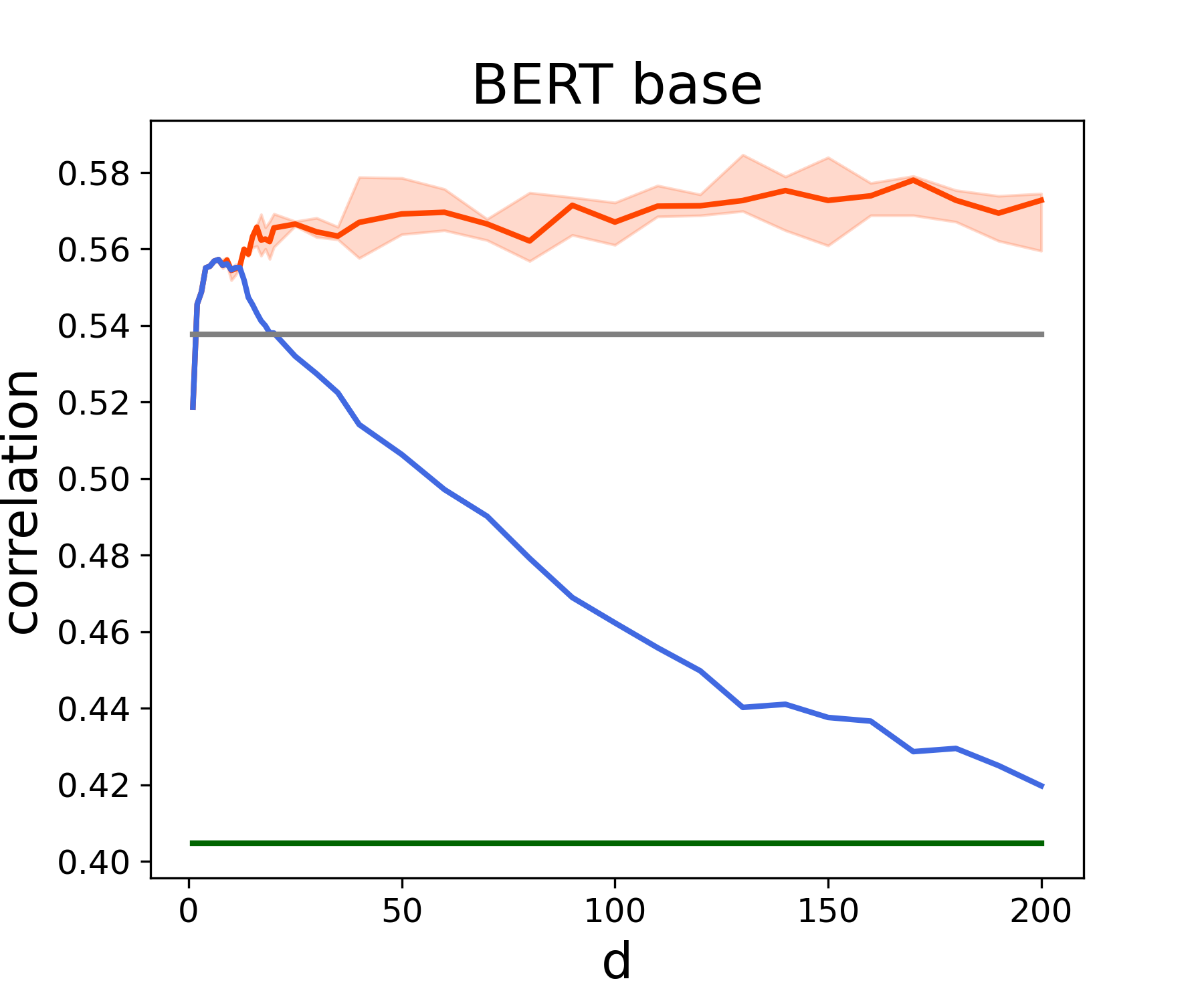}}\hfill
	\subcaptionbox{\label{fig3_b}}{\includegraphics[width=0.32\linewidth]{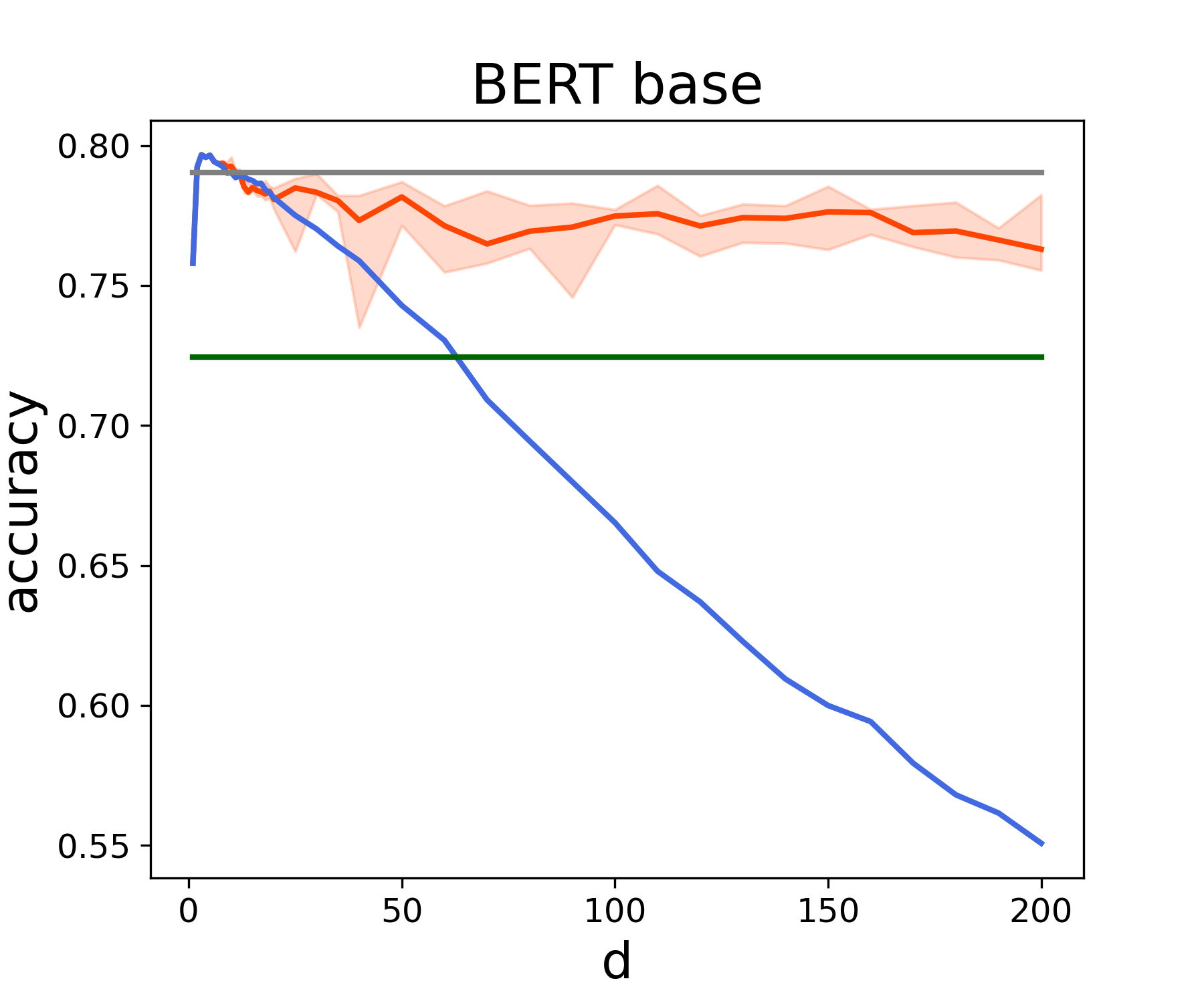}}\hfill
	\subcaptionbox{\label{fig3_c}}{\includegraphics[width=0.35\linewidth]{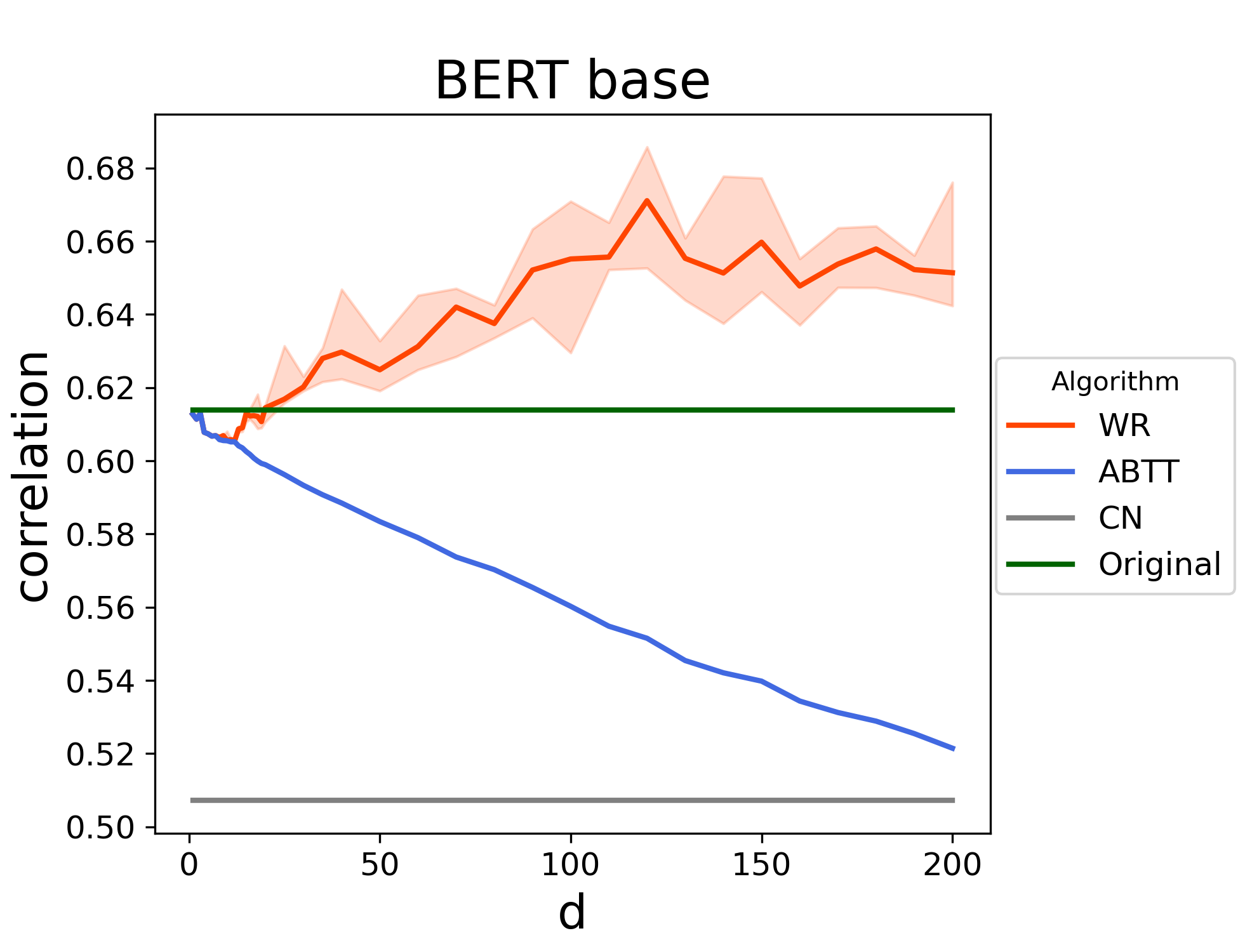}}\hfill

	\subcaptionbox{\label{fig3_d}}{\includegraphics[width=0.32\linewidth]{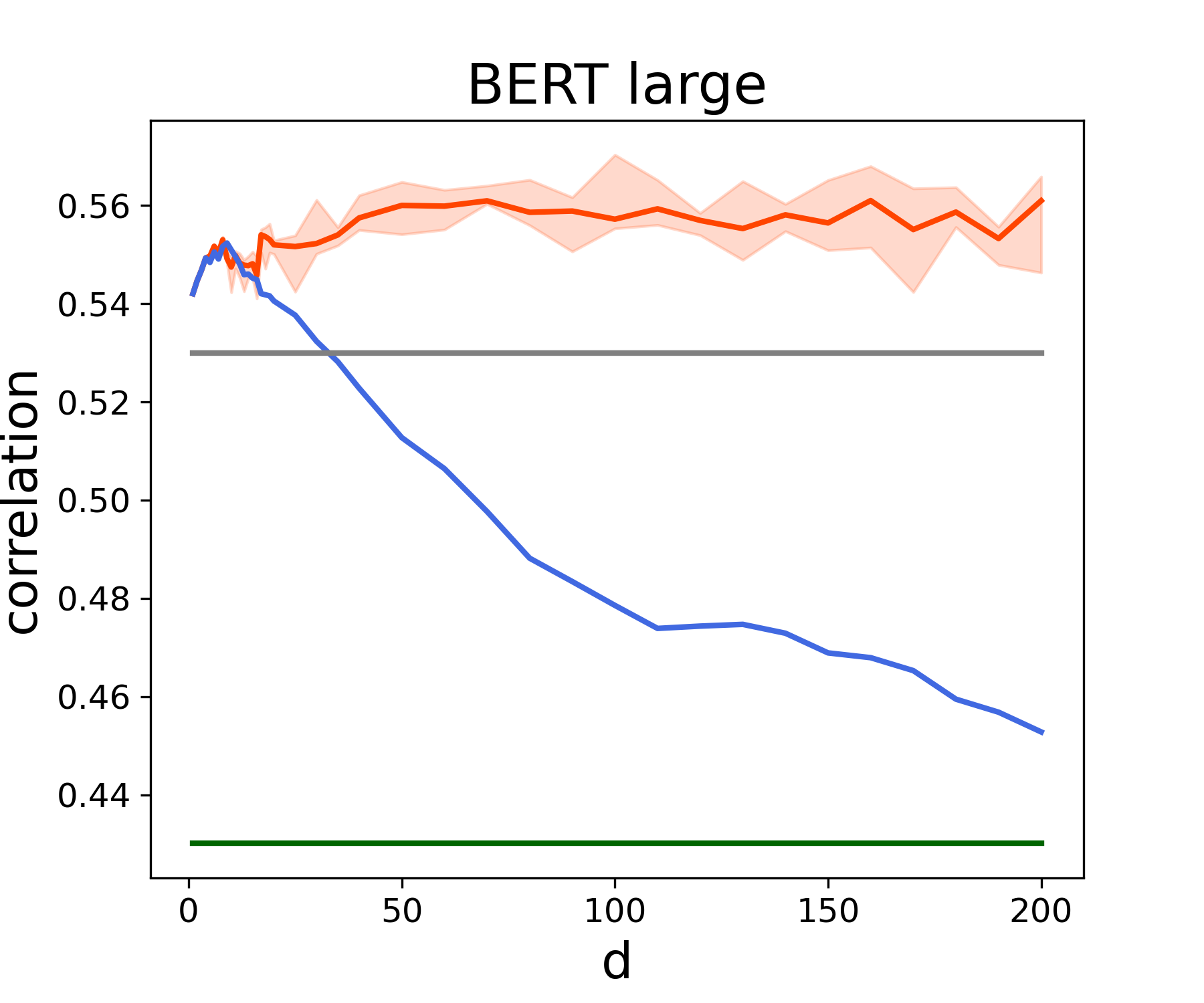}}\hfill	
	\subcaptionbox{\label{fig3_e}}{\includegraphics[width=0.32\linewidth]{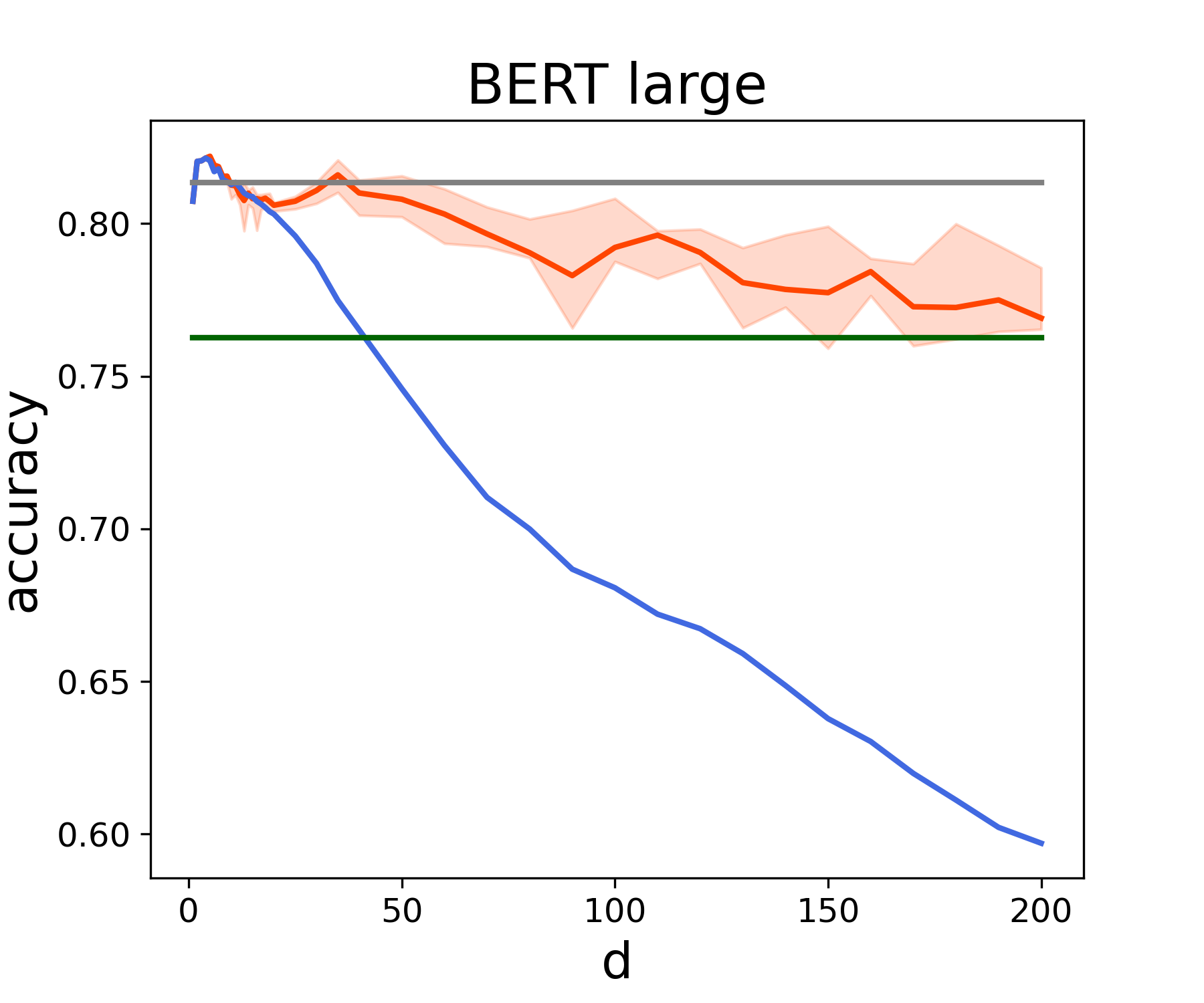}}\hfill
	\subcaptionbox{\label{fig3_f}}{\includegraphics[width=0.35\linewidth]{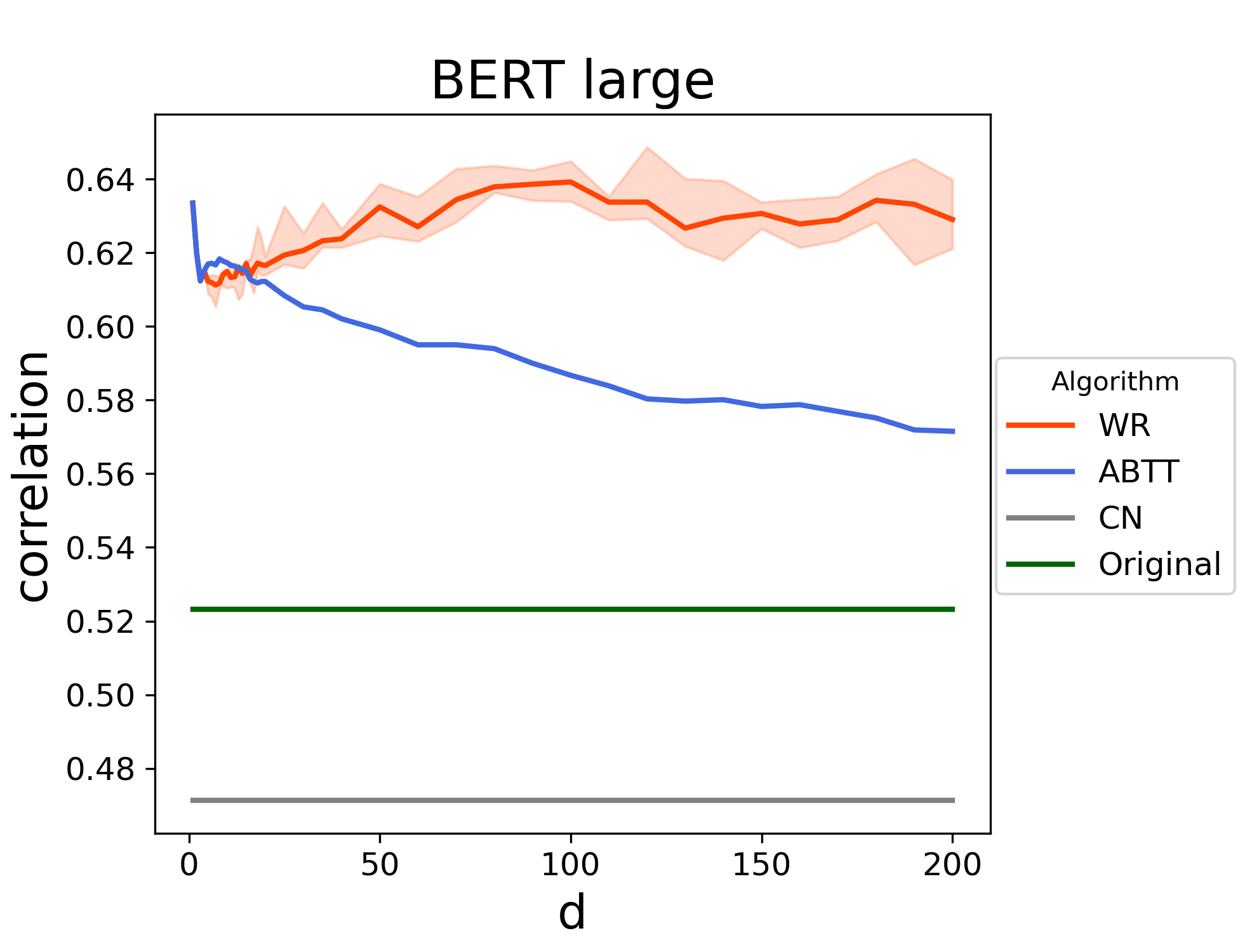}}\hfill
	\caption{Results of WR algorithm by different $d$ comparing to baselines: on BERT-base in (a) word similarity, (b) word analogy and (c) semantic textual similarity tasks; on BERT-large in (d) word similarity, (e) word analogy and (f) semantic textual similarity tasks.} \label{fig3}
\end{figure}

The overall results are shown in Figure \ref{fig3_a} and \ref{fig3_d}. As illustrated, the word embeddings processed by the WR algorithm are consistently and significantly outperform the original BERT embedding and CN algorithm. As the hyperparameter $d$ increases above 20, the correlation result of ABTT begins to drop, but the result of WR algorithm keeps stable and even rises slightly. The drop of ABTT's result could be caused by removing too many directions and losing expressiveness, and yet by adding weights to $d$ directions, WR algorithm could keep even improve the performance. The performance on eight word similarity datasets is provided in Table \ref{tab3}, note that ABTT and WR method achieve optimal results within the different ranges of $d$.

\begin{table*}[t]
	\caption{Correlation results on word similarity task.}
	\centering
	\label{tab3}
	\begin{tabular}{cllllllll}
		\hline
		\multirow{2}{*}{dataset} & \multicolumn{4}{c}{BERT-base}                              & \multicolumn{4}{c}{BERT-large}                                      \\ \cline{2-9} 
		& ORIG & CN      & ABTT          & WR                     & ORIG & CN      & ABTT                  & WR                      \\ \hline
		RG65      & .769  & .855 & .844(d=8)  & \textbf{.903(d=60)}  & .866  & .906& .900(d=1)           & \textbf{.934(d=100)}  \\
		SimVerb-3500  & .299  & .412 & .431(d=5)  & \textbf{.447(d=20)} & .310  & .416 & .452(d=1)           & \textbf{.456(d=110)} \\
		MEN-3000    & .517  & .673 & .702(d=6)  & \textbf{.722(d=50)}  & .562  & .663 & .700(d=6)  & \textbf{.709(d=100)}  \\
		Mturk287  & .370  & .716 & .747(d=6)  & \textbf{.768(d=160)} & .435  & .701 & \textbf{.723(d=4)}  & \textbf{.723(d=4)}  \\
		Mturk-771   & .529  & .652 & .667(d=5)  & \textbf{.676(d=60)}  & .564  & .646 & .666(d=3)           & \textbf{.669(d=6)}  \\
		RW-2034   & .493  & .665 & .668(d=9)  & \textbf{.707(d=200)} & .489  & .641 & .660(d=4)           & \textbf{.687(d=25)}  \\
		SimLex999   & .492  & .549 & .570(d=10) & \textbf{.581(d=200)} & .518  & .544 & .580(d=11)          & \textbf{.586(d=50)} \\
		WordSim353   & .569  & .637 & .665(d=10) & \textbf{.682(d=25)}  & .617  & .649 & .677(d=8)           & \textbf{.693(d=9)}   \\ \hline
	\end{tabular}
\end{table*}

\textbf{Word Analogy}: The purpose of word analogy task is to test the capability of a word embedding to encode linguistic relations between words. We use the analogy dataset introduced in \cite{mikolov2013efficient} containing 14 types of relations. These relations can be divided into two parts: the semantic relation such as capital of countries and the syntactic relation like adjective to adverb relation.

We calculate the word $w_4$ by finding the word in vocabulary that maximizes the cosine similarity between $v(w_4)$ and $v(w_2) - v(w_1) + v(w_3)$. The overall result of word analogy （(Figure \ref{fig3_b} and \ref{fig3_e}) shows that The  WR and ABTT methods both have a positive effect on the performance of word analogy task.

The best performance of each method on the word analogy task is provided in Table \ref{tab4}, we report the results by two types of relation: the semantic relation and the syntactic relation. The word vectors processed by WR perform better than the original ones with an average improvement of 13.04\%. 


\begin{table*}[t]
	\caption{Accuracy results on word analogy task.}
	\centering
	\label{tab4}
	\begin{tabular}{cllllllll}
		\hline
		\multirow{2}{*}{dataset} & 
		\multicolumn{4}{c}{BERT-base}  & \multicolumn{4}{c}{BERT-large}                                     \\ \cline{2-9} 
		& ORIG & CN& ABTT & WR& ORIG & CN & ABTT & WR\\ 
		\hline
		semantic & .655  & .768 & \textbf{.787(d=5)}    & \textbf{.787(d=5)}   & .688  & .775 & \textbf{.799(d=7)} & .797(d=7) \\
		syntactic                & .761  & .802 & .802(d=3)             & \textbf{.827(d=130)} & .802  & .834 & .835(d=2)          & \textbf{.863(d=90)} \\ 
		\hline
	\end{tabular}
\end{table*}

\textbf{Semantic Textual Similarity}: We also evaluate the WR algorithm's effect on downstream task like the semantic textual similarity task. The semantic textual similarity (STS for short) task is aimed to determine how similar two texts are. We use the STS Benchmark dataset which comprises a selection of datasets used in the STS tasks between 2012 and 2017.

We use a widely used and effective method to represent sentences: averaging the word embedding of each word in the sentence, namely, the sentence representation $v(s) = \frac{1}{|s|}\sum_{w \in s}v(w)$. The similarity between two sentences is calculated by the cosine similarity of their corresponding sentence representations. Pearson correlation between predictions and human judgments is used to evaluate an algorithm's performance.

The performances of WR and ABTT keep consistent when $d$ is smaller than 10. As $d$ exceeds a certain number, the performance of ABTT algorithm starts to drop, but the results of the WR algorithm keep growing and outperform all baselines, as shown in Figure \ref{fig3_c} and \ref{fig3_f}. The best performance on semantic textual similarity tasks is provided in Table \ref{tab5}, note that ABTT and WR method achieve optimal results within the different ranges of $d$. An average improvement of 16.78\% over the original word embedding suggests that the WR algorithm can significantly improve the performance of downstream tasks.

\begin{table*}[t]
	\caption{Correlation results on semantic textual similarity task.}
	\centering
	\label{tab5}
	\begin{tabular}{cllllllll}
		\hline
		\multirow{2}{*}{dataset} & \multicolumn{4}{c}{BERT-base}                                        & \multicolumn{4}{c}{BERT-large}                                       \\ \cline{2-9} 
		& ORIG & CN       & ABTT                  & WR                      & ORIG & CN       & ABTT                  & WR                      \\ \hline
		2012                     & .646 & .596 & .657(d=1)          & \textbf{.714(d=120)} & .588 & .588 & .663(d=1)          & \textbf{.684(d=120)} \\
		2013                     & .660 & .662 & .708(d=3)          & \textbf{.725(d=140)} & .639 & .640 & \textbf{.702(d=8)} & .690(d=200) \\
		2014                     & .562 & .477 & .578(d=3)          & \textbf{.634(d=120)} & .480 & .442 & .592(d=1)          & \textbf{.711(d=190)} \\
		2015                     & .639 & .516 & .657(d=3)          & \textbf{.702(d=150)} & .523 & .466 & .6687(d=8)          & \textbf{.683(d=80)} \\
		2016                     & .519 & .436 & .514(d=1)          & \textbf{.527(d=180)} & .441 & .408 & \textbf{.522(d=1)} & \textbf{.522(d=1)} \\
		2017                     & .616 & .577 & .682(d=3)          & \textbf{.727(d=150)} & .528 & .538 & .689(d=8)         & \textbf{.707(d=80)}  \\ \hline
	\end{tabular}
\end{table*}

\subsection{Towards Isotropy}
Our method also has an impact on the geometric characteristics of word embedding. In this section, we demonstrate the geometry of word embedding processed by the WR algorithm and make a comparison with the original one.

\begin{figure}[t]
	\centering
	\setlength{\lineskip}{\medskipamount}
	\subcaptionbox{\label{fig4_a}}{\includegraphics[width=0.46\linewidth]{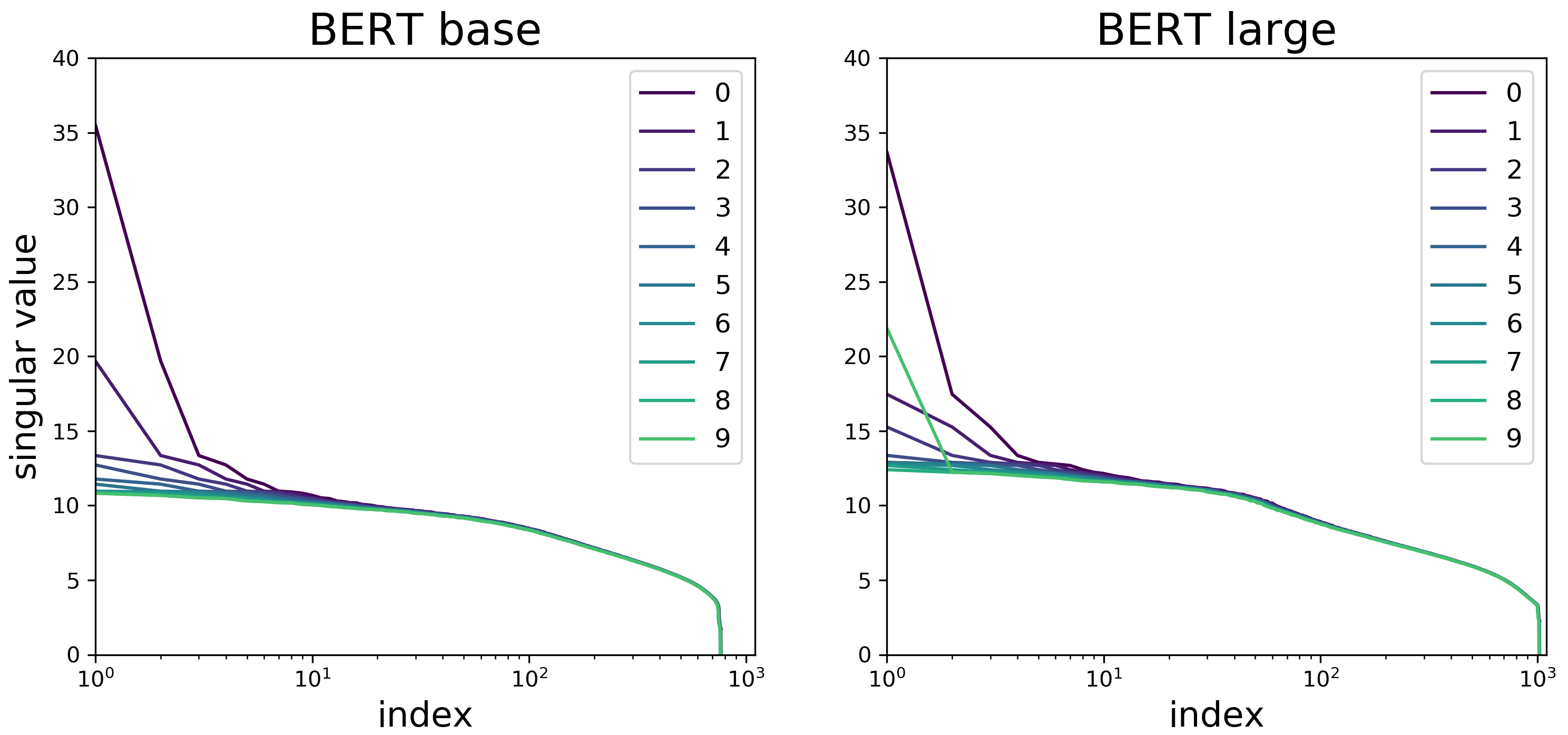}}\hfill
	\subcaptionbox{\label{fig4_b}}{\includegraphics[width=0.54\linewidth]{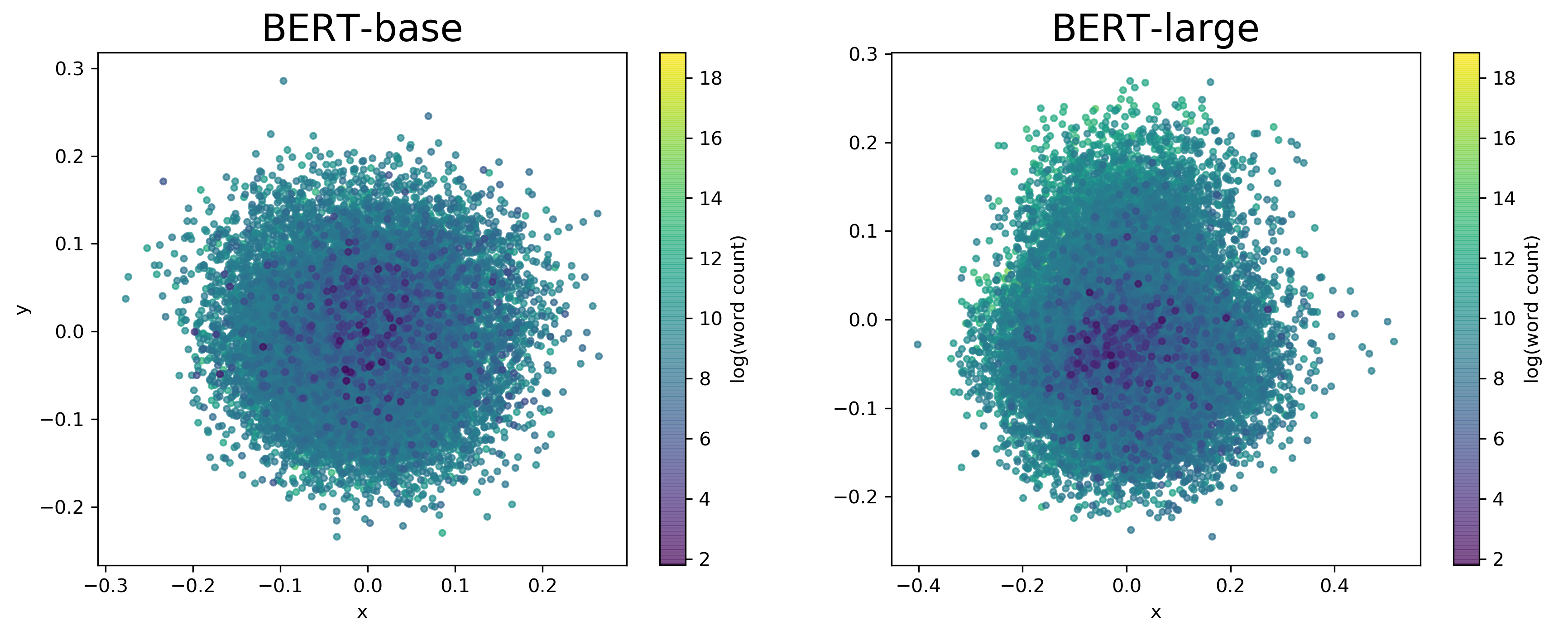}}\hfill
	\caption{(a) Singular values of embedding matrices processed by WR with different $d$. (b) Projecting word embeddings (processed by WR when $d$ equals 5) onto top 2 PCA directions and use color to indicate word frequency.} \label{fig4}
\end{figure}

Our method can fix the fast singular value decay problem. After applying the WR algorithm, the first several singular values decrease significantly which shows reducing of the imbalance between different directions in the word embedding. As shown in Figure \ref{fig4_a}, singular values are steady across directions after removing several dominant directions.

Figure \ref{fig4_b} visualizes processed BERT embedding by projecting onto the top two PCA directions and use color to indicate word frequency. The processed embedding is centered around the origin and words with different frequency are uniformly distributed which avoid the disproportionate impact of word frequency on the entire embedding. The average cosine similarity of words also drops to near zero after applying the WR algorithm which indicates the isotropic geometry of the processed word embedding.

\subsection{Discussion}
As the number of directions to be removed increases, we find a similar pattern in the performances of ABTT and WR algorithm in three evaluation tasks. When $d$ is small, the results of ABTT and WR keep consistent and the weights that the WR algorithm learned are close to 1. That means for the first several dominant directions, complete removal is optimal.

As $d$ grows above a certain number, the performance of ABTT starts to drop while WR's results keep steady in general and even increase. This demonstrates the ability of the WR method to maintain useful information in it by acquiring a weight for each direction.


\section{Conclusion}
In this paper, we show that the pre-trained BERT embedding is anisotropy which hurts its expressiveness. To correct the anisotropy, we propose a method to remove several dominant directions computed by PCA with a learnable weight to each direction. We train the weights on word similarity tasks and evaluate our method on three tasks: word similarity, word analogy, and semantic textual similarity. We compare our weighted removal method with three baselines and our method outperforms the baselines in most conditions.

There are many possibilities in this direction: other single factors that have a disproportionate impact on word embeddings, changing the training task of our method to obtain other improvements, combining our method with the transformer architecture, etc. We hope to investigate these issues in future work.

\section*{Acknowledgment}
\&
Our work is supported by the National Key Research and Development Program of China under grant No.2019YFC1521400 and National Natural Science Foundation of China under grant No.62072362.

%
%
%
%
%
\bibliographystyle{splncs04}
\bibliography{./sample-base}
%
\end{document}